\documentclass[letterpaper]{article} % DO NOT CHANGE THIS
\usepackage[preprint]{aaai2027}  % DO NOT CHANGE THIS
\usepackage[hyphens]{url}  % DO NOT CHANGE THIS
\usepackage{graphicx} % DO NOT CHANGE THIS
\usepackage{natbib}  % DO NOT CHANGE THIS AND DO NOT ADD ANY OPTIONS TO IT
\usepackage{amsmath}
\usepackage{caption} % DO NOT CHANGE THIS AND DO NOT ADD ANY OPTIONS TO IT
\usepackage{algorithm}
\usepackage{algorithmic}

\usepackage{multirow}
\usepackage{makecell}
\usepackage{xcolor}
\usepackage{graphicx}
\newcommand{\cmark}{\ding{51}}

\usepackage[table]{xcolor}
\definecolor{bestred}{RGB}{255,205,205}
\newcommand{\best}[1]{\begingroup\setlength{\fboxsep}{1.2pt}\colorbox{bestred}{\strut\textbf{#1}}\endgroup}
\usepackage{pifont}

\DeclareRobustCommand{\best}[1]{%
  \begingroup
  \setlength{\fboxsep}{1.2pt}%
  \colorbox{bestred}{\strut\textbf{#1}}%
  \endgroup
}

\DeclareRobustCommand{\bestcap}[1]{%
  \begingroup
  \setlength{\fboxsep}{1.2pt}%
  \colorbox{bestred}{\strut\textbf{#1}}%
  \endgroup
}

\newcommand{\xmark}{\textcolor{red!75!black}{\ding{55}}}

\usepackage{newfloat}
\usepackage{listings}
\DeclareCaptionStyle{ruled}{labelfont=normalfont,labelsep=colon,strut=off} % DO NOT CHANGE THIS
\floatstyle{ruled}
\newfloat{listing}{tb}{lst}{}
\floatname{listing}{Listing}

\usepackage{booktabs}

\title{DeliCIR: Memory-Guided Test-Time Deliberation via Multi-Agent Collaboration for Composed Image Retrieval}
\author{Xingtian Pei\textsuperscript{*},
Yukun Song\textsuperscript{*},
Changwei Wang,
Shunpeng Chen,\\
Rongtao Xu,
Shengpeng Xu,
Shibiao Xu\textsuperscript{$\dagger$} 
}
\affiliations{
Beijing University of Posts and Telecommunications\\
Beijing, China\\
\textsuperscript{*}Equal contribution.\\
\textsuperscript{$\dagger$}Corresponding author: \texttt{shibiaoxu@bupt.edu.cn}.
}

\begin{document}

\maketitle

\begin{abstract}
Composed Image Retrieval (CIR) requires both preserving the visual continuity of the reference image and faithfully executing the semantic variables specified in the modification text, which constitute the core challenge of the task. Existing methods often suffer from Perception Myopia in a single space, or fall into Logic Drift in iterative collaboration due to the perception ceiling of the underlying retriever. To address this issue, we propose a one-stop hierarchical Perception-to-Deliberation Framework (PDF), which, to the best of our knowledge, is the first to introduce Test-Time Scaling (TTS) into CIR. Relying on a hierarchical multi-agent architecture, PDF first utilizes an Intent Routing Manager to dynamically dispatch multi-view Worker perception signals based on modification intents to construct a high-recall candidate pool. Subsequently, the Deliberation Manager combines a training-free reasoning experience distillation mechanism with a tournament-style test-time scaling strategy (T-TTS) to perform experience-guided fine-grained reasoning and produce the final retrieval results. Experimental results demonstrate that PDF achieves SOTA performance on three benchmark datasets: CIRR, CIRCO, and Fashion IQ. This study indicates that experience-driven memory mechanism and TTS represent a highly promising and scalable path for achieving fine-grained multimedia retrieval. The code will be made publicly available upon acceptance.
\end{abstract}


\section{Introduction}

\begin{figure}[!t]
    \centering
    \includegraphics[width=\linewidth]{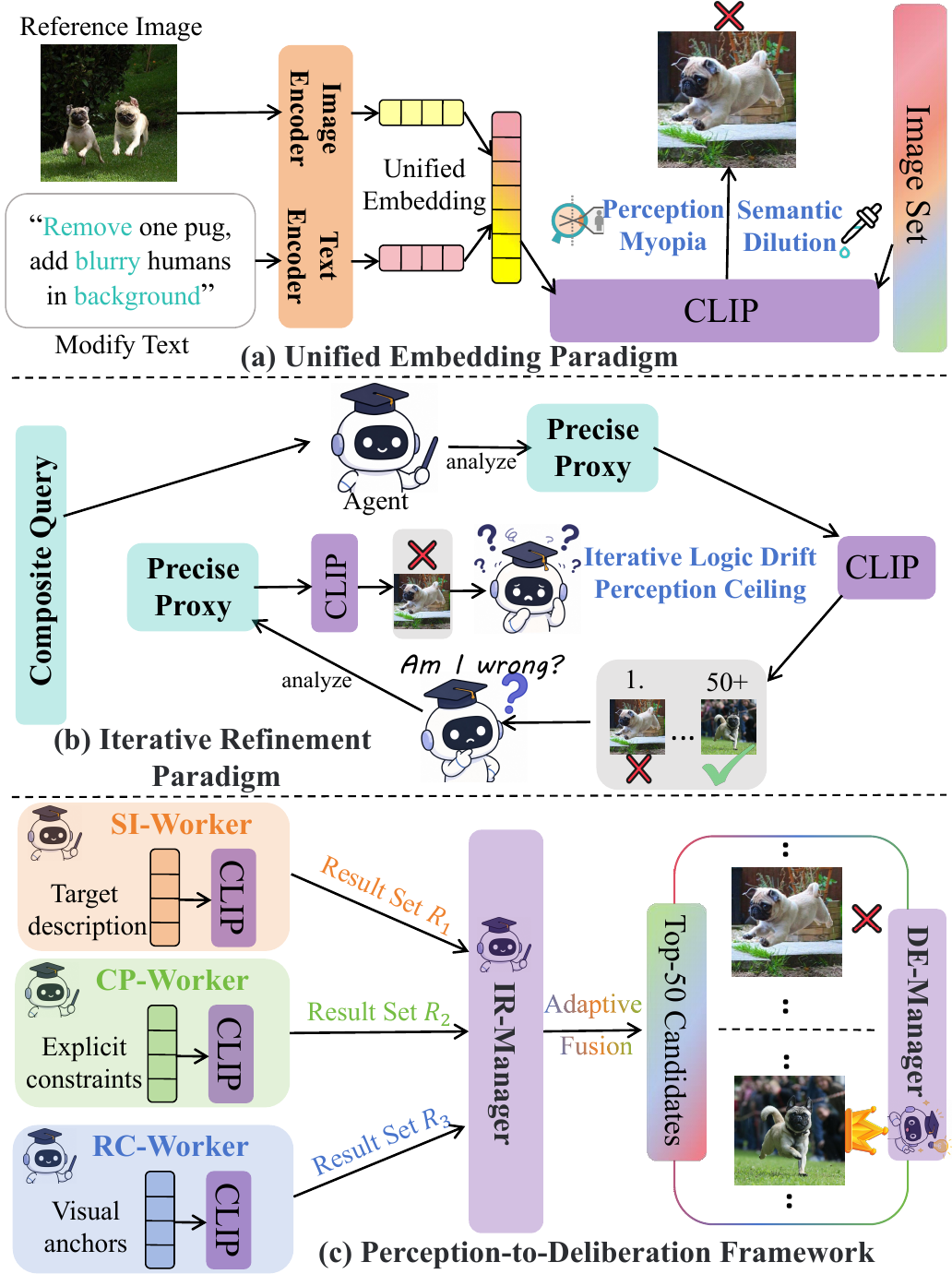}
    \caption{Comparison of three mainstream CIR paradigms.
    (a) Unified embedding methods suffer from perceptual myopia and semantic dilution.
    (b) Iterative refinement methods are constrained by logical drift and perception ceilings.
    (c) Efficient one-stop discriminative retrieval framework (Ours).
    }
    \label{fig:compare}
    \vspace{-1.0em}
\end{figure}

CIR allows users to express visual revision intents through combinations of reference images and modification texts, greatly facilitating human-computer interaction~\cite{song2025comprehensive,du2025survey}. Consequently, this retrieval task boasts broad application prospects in diverse scenarios including e-commerce~\cite{goenka2022fashionvlp,chen2025mai}, interactive image editing~\cite{brooks2023instructpix2pix,hertz2022prompt}, large-scale multimedia navigation~\cite{ventura2024covr}, and search engines~\cite{gu2021image}. Unlike traditional image-text matching tasks, the reference image in CIR serves as a visual continuity anchor rather than a complete representation of the retrieval intent, while the modification text consists of instruction-oriented text with operational tendencies. This requires the model to be able to parse composite instructions containing compositional descriptions and multiple explicit constraints~\cite{tang2024manipulation,sun2025leveraging}. The core challenge of CIR lies in moving beyond superficial feature alignment toward deep cross-modal reasoning that integrates heterogeneous signals.

Currently, CIR primarily follows two paradigms, as shown in Fig.~\ref{fig:compare}(a) and \ref{fig:compare}(b): The first paradigm relies on unified embedding models to map multimodal queries into a shared space~\cite{vo2019composing, baldrati2022conditioned, baldrati2023composed}, and is often constrained by perceptual myopia and semantic dilution: a single similarity metric struggles to accommodate diverse modification intents, while simple feature aggregation may cause the modification instructions to be overwhelmed by the original visual information, leading to alignment deviations in complex scenarios. The second paradigm introduces agentic frameworks that leverage large language models for iterative retrieval~\cite{cheng2025generative, tu2025multimodal}. Despite their flexibility, these methods face two key bottlenecks: First, they mostly adopt static flat coordination architectures, lacking hierarchical orchestration to balance conflicting priorities, which makes logical drift difficult to avoid during multi-round feedback; second, by using basic retrievers as the final retrieval basis, the high-order semantic increments produced by LLMs are often diluted by the coarse representation of the underlying retrievers. This creates a structural imbalance where logic leads but perception lags, causing the final performance to be bounded by the perception ceiling of the underlying retriever.

Based on the above insights, we propose the Perception-to-Deliberation Framework (PDF) for deliberate reasoning, as shown in Fig.~\ref{fig:compare}(c). To overcome the perceptual myopia caused by single-similarity retrieval, PDF deploys three Worker Agents to extract target imagination, explicit constraints, and visual anchors, thereby replacing the conventional single-metric paradigm with a multi-view prior system. To address the static and flat coordination bottleneck of existing agentic paradigms, PDF constructs a dynamic hierarchical collaboration architecture, where the Intent Routing Manager (IR-Manager) adaptively schedules perceptual resources according to query intent. Furthermore, to break through the perceptual ceiling imposed by the underlying retriever, we propose Reasoning Experience Distillation (RED) and the Tournament-based Test-Time Scaling (T-TTS) mechanism, which reshape the retriever into a lower-bound provider for high-recall retrieval, while treating high-density deliberate refinement as the upper bound determining discriminative precision. By systematically increasing the reasoning depth, this one-stop decision flow achieves logical self-verification and thereby overcomes the perceptual bottleneck of pretrained models without requiring repeated retrieval iterations. In summary, the main contributions of this work are as follows:
\begin{itemize}
    \item \textbf{Hierarchical Multi-agent Collaboration Architecture:} We propose a Worker-Manager hierarchical system that deconstructs multi-view perception, intent routing, and deliberate decision-making into specialized layers, achieving more strategic multimodal information synergy compared to existing flat agent architectures.
    \item \textbf{Intent-aware Prior Routing (IPR):} We introduce a dynamic fusion mechanism that adaptively assigns weights to multi-view retrieval priors according to specific query intents. This effectively mitigates the semantic drift caused by over-reliance on a single retrieval branch.
    \item \textbf{Memory-Guided Reasoning Refiner (MRR):} To the best of our knowledge, we are the first to enhance fine-grained candidate discrimination for complex compositional queries in CIR through an experience-driven test-time scaling strategy, effectively converting additional inference compute into retrieval accuracy gains.
\end{itemize}

\section{Related Work}
\subsection{Composed Image Retrieval}
Supervised composed image retrieval relies on costly triplet annotations to accomplish target image matching~\cite{cohen2022my, li2025learning, feng2025vqa4cir, gao2025ncl, chen2026intent}. To reduce the dependence on annotated data, zero-shot CIR methods leverage the prior knowledge of pretrained vision-language models to alleviate cross-modal asymmetry~\cite{saito2023pic2word, lin2024fine, gu2024language, li2025hierarchy, li2025rethinking}. Subsequently, some works introduce knowledge distillation~\cite{baldrati2023zero}, diffusion models, and synthetic data~\cite{gu2023compodiff, xing2025context, wang2025generative} to enhance cross-modal alignment, but they still face limitations in cross-domain generalization. With the development of large language models, training-free CIR methods exploit the generation capabilities~\cite{yang2024ldre, li2025imagine} and reasoning capabilities~\cite{karthik2024vision, sun2025cotmr, luo2025imagescope, tang2025reason} of LLMs to bridge semantic gaps. However, such flat paradigms remain limited in correcting the perceptual biases inherited from the underlying retrievers. Furthermore, MRA-CIR~\cite{tu2025multimodal} explicitly decouples intent understanding from execution, AutoCIR~\cite{cheng2025generative} introduces a corrective feedback loop to enhance system adaptivity, and $X^{R}$~\cite{yang2026xr} incorporates cross-modal collaborative planning for multi-stage filtering. Despite their improved performance, these methods still largely rely on heuristic and static orchestration, making it difficult to adaptively allocate reasoning resources for different query intents.

\begin{figure*}[!t]
    \centering
    \includegraphics[width=\textwidth]{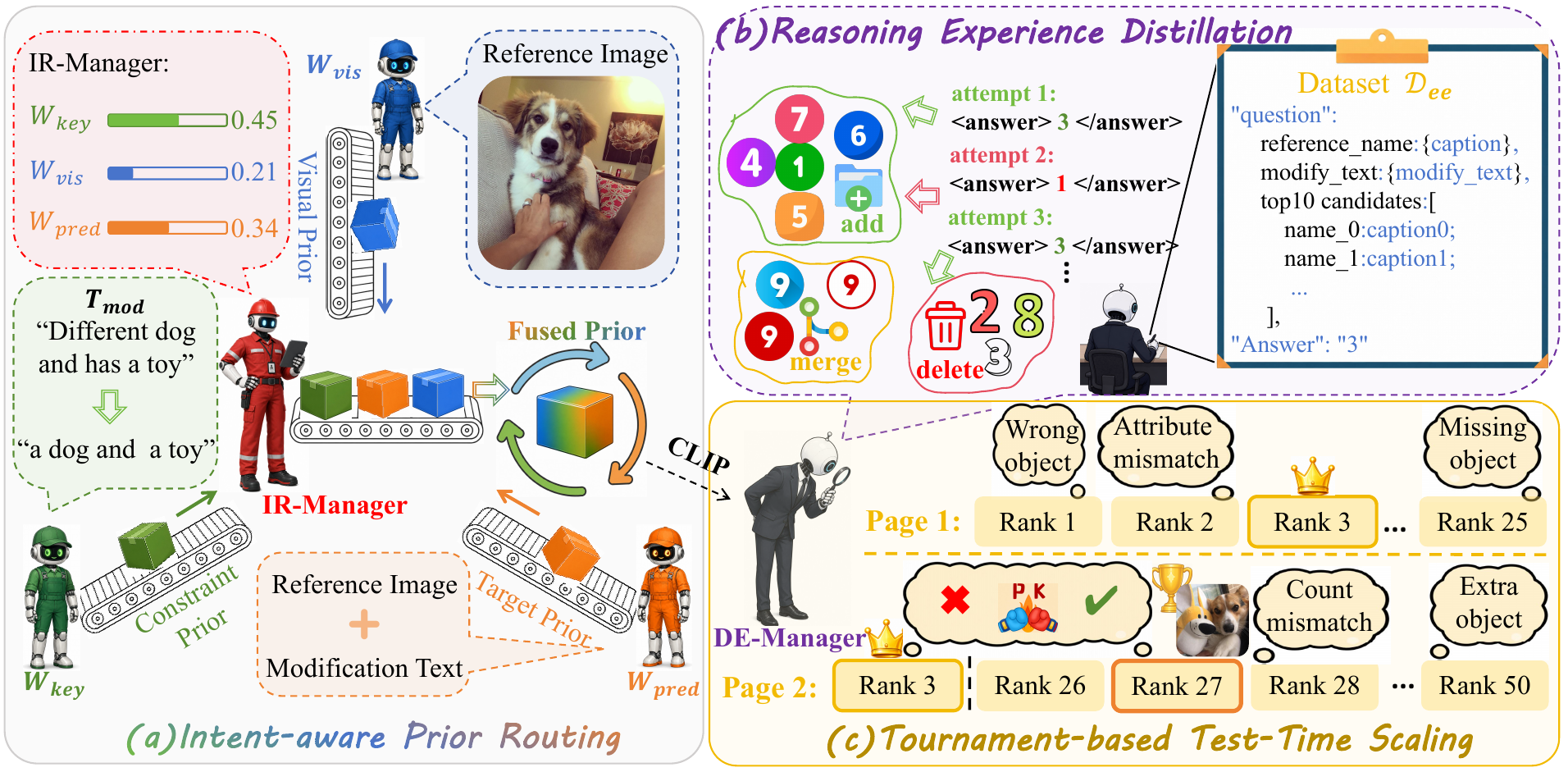}
    \caption{
    Overview of the proposed PDF.
    (a) IPR adaptively fuses multi-view priors to construct a high-quality candidate pool.
    (b) RED consolidates the heuristics distilled from offline deliberation into an experience memory.
    (c) The T-TTS strategy performs page-wise competitive screening over the candidate pool, converting test-time computation into improved semantic alignment.
    }
    \label{fig:main_framework}
\end{figure*}

\subsection{Agentic Reasoning and Test-Time Scaling}
Multi-agent systems (MAS) have demonstrated strong collaborative capabilities in multimedia processing and interpretable reasoning~\cite{lan2023collaborative, lin2023local, wang2026towards, li2025signed}. As AI agents evolve from static generation toward autonomous planning, memory has gradually become a key foundation that connects perception, reasoning, and self-evolution~\cite{hu2025memory}. External memory mechanisms provide experience retention and knowledge reuse capabilities for complex decision-making, thereby helping improve system intelligence~\cite{peng2025bmb, cheng2025distribution, yu2025mossvln, yu2025vismem}. In the CIR domain, MAI~\cite{chen2025mai} is among the early works to introduce memory modules to alleviate semantic forgetting and information redundancy in multi-turn interactions. On the other hand, Test-Time Scaling (TTS) improves reasoning and decision-making capabilities by increasing the test-time computational budget~\cite{wei2022chain, wang2022self, yao2023tree,openai2024learning, snell2025scaling}, and has achieved promising results in tasks such as code generation and multimodal understanding~\cite{li2025s, yu2025z1, muennighoff2025s1, liu2025video, tian2026unigen}. It has further inspired various test-time learning mechanisms~\cite{suzgun2026dynamic, xiao2025dynaprompt, hu2025test, zhao2026genprm, weller2025rank1, yang2026towards}. Although TT-RLDR has attempted to introduce TTS into CIR, its core still relies on parameter optimization during the test stage~\cite{zhou2026duplex}. In contrast, PDF requires no parameter updates. Instead, it converts test-time computation into deliberate candidate comparison and rank correction through hierarchical agentic collaboration, experience-driven memory, and tournament-style TTS.

\section{Methodology}
\subsection{Problem Formulation and Framework Overview}
\textbf{Task Definition.} CIR aims to retrieve target images $I_{tar}$ satisfying the modification intent from a large-scale gallery  $\mathcal{G}=\{I_1,I_2,\dots,I_N\}$ based on a composed query $q=(I_{ref},T_{mod})$. Here, $I_{ref}$ denotes the reference image and $T_{mod}$ denotes the modification text. The ground-truth set corresponding to query $q$ is denoted as $\mathcal{Y}_q \subseteq \mathcal{G}$. For single-ground-truth datasets such as CIRR and Fashion IQ, we have $|\mathcal{Y}_q|=1$; while for the multi-ground-truth dataset CIRCO, we have $|\mathcal{Y}_q|>1$.

\textbf{Visual Semantic Proxy Construction.} In CIR, $T_{mod}$ is an instruction-oriented text with operational tendencies. It usually omits the subject or contextual anchors, yet implicitly imposes complex executable constraints on the content of $I_{ref}$. Such a query form requires the model not only to understand the semantics of $T_{mod}$, but also to extract sufficient and complementary visual semantics from $I_{ref}$. Therefore, we construct a visual semantic proxy for $I \in \mathcal{G}$ as follows:
\begin{equation}
C(I) = \Phi_{vsp}(I),
\end{equation}
where $\Phi_{vsp}(\cdot)$ represents the visual semantic proxy constructor, and $C(I)$ is a dense semantic transcription of $I$. This representation preserves visual anchors as evidence for subsequent deliberate reasoning.

\textbf{Overview.} Our PDF framework, as shown in Fig.~\ref{fig:main_framework}, first employs the Semantic Imagination Worker Agent (SI-Worker), Constraint Parsing Worker Agent (CP-Worker), and Reference Consistency Worker Agent (RC-Worker) to extract priors from three complementary perspectives and generate candidate results. Subsequently, the Intent Routing Manager agent (IR-Manager) adaptively fuses these three results according to the prior intent of $T_{mod}$, thereby constructing a high-recall candidate pool $\mathcal{B}_K$ (Sec.~3.2). The Deliberation Experience Manager agent (DE-Manager) then performs high-precision candidate discrimination within $\mathcal{B}_K$ based on the \(C(I)\) of candidate images through memory-enhanced T-TTS (Sec.~3.3), and directly outputs the final retrieval results.

\subsection{Intent-Aware Weighted Rank Fusion}
\textbf{Multi-View Perception Workers.} The SI-Worker utilizes a semantic imagination operator $W_{pred}$ to fuse the semantic proxy $C(I_{ref})$ of the reference image with $T_{mod}$ into a target semantic hypothesis $C_{pred} = W_{pred}(C(I_{ref}), T_{mod})$. This hypothesis characterizes the potential target semantics under the influence of the modification intent. Retrieval is then performed using the CLIP text encoder $\phi_t(\cdot)$ and visual encoder $\phi_v(\cdot)$:
\begin{equation}
R_{pred} = \text{Rank}_{I \in \mathcal{G}} \left( \text{sim} \left( \phi_t(C_{pred}), \phi_v(I) \right) \right).
\end{equation}

The CP-Worker extracts explicit visual constraints that must be satisfied from the modification text: $\mathcal{K} = W_{key}(T_{mod}) = \{ (o_t, a_t) \}_{t=1}^M$, where $W_{key}$ is a constraint extraction operator, and $o_t, a_t$ represent the key visual objects and their attributes directly related to the target content. Through a declarative mapping function $\Psi(\cdot)$, imperative structures and comparative terms are normalized into a compact declarative description $C_{key} = \Psi(\mathcal{K})$ suitable for text-based retrieval. The resulting modification-semantic ranking is:
\begin{equation}
R_{key} = \text{Rank}_{I \in \mathcal{G}} \left( \text{sim} \left( \phi_t(C_{key}), \phi_v(I) \right) \right).
\end{equation}

The RC-Worker uses the reference image as a visual anchor. By measuring the distance between $I$ and $I_{ref}$ in the original feature manifold, it anchors the visual tone of the retrieval process in a high-dimensional feature space to prevent semantic drift caused by over-editing. The visual anchor ranking is obtained as:
\begin{equation}
R_{vis} = \text{Rank}_{I \in \mathcal{G}} \left( \text{sim} \left( \phi_v(I_{ref}), \phi_v(I) \right) \right).
\end{equation}
Through this multi-view perception, we obtain three complementary candidate rankings $\{R_{pred}, R_{key}, R_{vis}\}$ that capture target evolution logic, attribute hard constraints, and visual ancestry, respectively.

\textbf{Dynamic Intent Routing and Weighted Rank Fusion.} Given the three candidate rankings, we further introduce the IR-Manager $\mathcal{M}_{route}(\cdot)$ to model the semantic properties of $T_{mod}$ and adaptively assign an intent-aware weight vector:
\begin{equation}
w = \mathcal{M}_{route}(T_{mod}) = [w_m], \ w_m \ge 0, \sum_m w_m = 1,
\end{equation}
where $m \in \{\text{pred}, \text{key}, \text{vis}\}$ denotes the index of each Worker Agent. This weight allocation reflects the varying dependencies of different queries on the three types of priors: when the modification intent leans toward holistic semantic reconstruction, $w_{pred}$ is relatively increased; when the modification intent can form complete explicit constraints, $w_{key}$ is relatively increased; and when the modification intent requires preserving the visual tone of $I_{ref}$, $w_{vis}$ should be strengthened. We then perform weighted fusion at the ranking level to enhance cross-branch robustness. Let $r_m(I)$ be the rank of image $I \in \mathcal{G}$ in the $m$-th perspective, the fused score is calculated as:
\begin{equation}
S_{fuse}(I|q) = \sum_{m} w_m \cdot \frac{1}{r_m(I) + \tau},
\end{equation}
where $\tau > 0$ is a smoothing constant used to regulate the attenuation rate of reciprocal ranks. By suppressing the weight dominance of top-ranked candidates, it encourages consensus-seeking across multiple views and enhances robustness against local noise and ranking fluctuations. Finally, the candidate set $\mathcal{G}$ is re-ranked according to $S_{fuse}(I|q)$, and the top-$K$ candidates are truncated to construct the high-recall candidate pool:
\begin{equation}
\mathcal{B}_K = \{ I_{(1)}, I_{(2)}, \dots, I_{(K)} \}.
\end{equation}

\begin{table*}[t]
\centering

% --- 在表格上方定义这些宏命令，确保格式统一 ---
% 1. 定义会议号缩放逻辑
\newcommand{\conf}[1]{\hspace{1pt}\scalebox{0.75}{\textit{#1}}}

% 2. 定义符合你要求的 method 宏：参数为 {名称}{引用Key}{会议名}
\newcommand{\methoditem}[3]{#1~\cite{#2}~\conf{#3}}

% 3. 定义专门缩小的 XR 上标
\newcommand{\XR}{X\textsuperscript{\scalebox{0.6}{R}}}
% -------------------------------------------

\caption[Comparison with representative baselines on CIRCO and CIRR.]{
\textbf{Comparison with representative baselines on the test sets of CIRCO and CIRR.}
The best results are highlighted with \bestcap{red boxes}, while the second-best results are \underline{underlined}.
``--'' indicates results not reported in the original paper.
}
\label{tab:circo_cirr_comparison}
\scriptsize 
\setlength{\tabcolsep}{2.1pt}
\renewcommand{\arraystretch}{1.1}
\resizebox{\textwidth}{!}{
\begin{tabular}{lllc|cccc|cccc|ccc}
\Xhline{0.8pt}

\multirow{3}{*}{\centering\textbf{Backbone}}
& \multirow{3}{*}{\centering\textbf{Type}}
& \multirow{3}{*}{\centering\textbf{Method}}
& \multirow{3}{*}{\makecell{\textbf{Training-}\\\textbf{free}}}
& \multicolumn{4}{c|}{\textbf{CIRCO}}
& \multicolumn{7}{c}{\textbf{CIRR}} \\

\cline{5-15}

& & &
& \multicolumn{4}{c|}{mAP@$K$}
& \multicolumn{4}{c|}{Recall@$K$}
& \multicolumn{3}{c}{$R_s$@$K$} \\

& & &
& $k=5$ & $k=10$ & $k=25$ & $k=50$
& $k=1$ & $k=5$ & $k=10$ & $k=50$
& $k=1$ & $k=2$ & $k=3$ \\

\hline

\multirow{10}{*}{\centering ViT-B/32}

& \multirow{7}{*}{\centering\makecell{CIR-\\dedicated}}

& \methoditem{PALAVRA}{cohen2022my}{ECCV'22} & \xmark 
& 4.61 & 5.32 & 6.33 & 6.80 
& 16.62 & 43.49 & 58.51 & 83.95 
& 41.61 & 65.30 & 80.94 \\

& & \methoditem{SEARLE}{baldrati2023zero}{ICCV'23} & \xmark 
& 9.35 & 9.94 & 11.13 & 11.84 
& 24.00 & 53.42 & 66.82 & 89.78 
& 54.89 & 76.60 & 88.19 \\

\cline{3-15}

& & \methoditem{CIReVL}{karthik2024vision}{ICLR'24} & \cmark
& 14.94 & 15.42 & 17.00 & 17.82 
& 23.94 & 52.51 & 66.00 & 86.95 
& 60.17 & 80.05 & 90.19 \\

& & \methoditem{LDRE}{yang2024ldre}{SIGIR'24} & \cmark
& 17.96 & 18.32 & 20.21 & 21.11 
& 25.69 & 55.13 & 69.04 & 89.90 
& 60.53 & 80.65 & 90.70 \\

& & \methoditem{ImageScope}{luo2025imagescope}{WWW'25} & \cmark
& 22.36 & 22.19 & 23.03 & 23.83 
& 34.36 & 60.58 & 71.40 & 88.41 
& 74.63 & 87.93 & 93.83 \\

& & \methoditem{OSrCIR}{tang2025reason}{CVPR'25} & \cmark
& 18.04 & 19.17 & 20.94 & 21.85 
& 25.42 & 54.54 & 68.19 & -- 
& 62.31 & 80.86 & 91.13 \\

& & \methoditem{CoTMR}{sun2025cotmr}{ICCV'25} & \cmark
& 22.23 & 22.78 & 24.68 & \underline{25.74} 
& 31.50 & 60.80 & 73.04 & 91.06 
& 66.61 & 84.50 & 92.55 \\

\cline{2-15}

& \multirow{3}{*}{\centering Agentic}

& \methoditem{AutoCIR}{cheng2025generative}{KDD'25} & \cmark
& 18.82 & 19.41 & 21.38 & 22.32 
& 30.53 & 59.42 & 72.19 & 91.47 
& 65.11 & 84.02 & 92.70 \\

& & \methoditem{$\mathrm{X}^{\mathrm{R}}$}{yang2026xr}{WWW'26} & \cmark
& \underline{27.51} & \underline{28.33} & \underline{30.28} & \underline{30.95} 
& \underline{43.06} & \underline{73.86} & \best{83.15} & \underline{94.36} 
& \underline{77.54} & \underline{90.27} & \best{95.21} \\

& & \textbf{PDF (Ours)} & \cmark
& \best{40.02} & \best{40.12} & \best{41.75} & \best{42.50} 
& \best{43.33} & \best{74.90} & \underline{82.88} & \best{94.80} 
& \best{78.66} & \best{90.41} & \underline{94.70} \\

\Xhline{0.8pt}

\multirow{14}{*}{\centering ViT-L/14}

& \multirow{10}{*}{\centering\makecell{CIR-\\dedicated}}

& \methoditem{Pic2Word}{saito2023pic2word}{CVPR'23} & \xmark
& 8.72 & 9.51 & 10.64 & 11.29 
& 23.90 & 51.70 & 65.30 & 87.80 
& -- & -- & -- \\

& & \methoditem{SEARLE}{baldrati2023zero}{ICCV'23} & \xmark
& 11.68 & 12.73 & 14.33 & 15.12 
& 24.24 & 52.48 & 66.29 & 88.84 
& 53.76 & 75.01 & 88.19 \\

& & \methoditem{LinCIR}{gu2024language}{CVPR'24} & \xmark
& 12.59 & 13.58 & 15.00 & 15.85 
& 25.04 & 53.25 & 66.68 & -- 
& 57.11 & 77.37 & 88.89 \\

& & \methoditem{FTI4CIR}{lin2024fine}{SIGIR'24} & \xmark
& 15.05 & 16.32 & 18.06 & 19.05 
& 25.90 & 55.61 & 67.66 & 89.66 
& 55.21 & 75.88 & 87.98 \\

\cline{3-15}

& & \methoditem{CIReVL}{karthik2024vision}{ICLR'24} & \cmark
& 18.57 & 19.01 & 20.89 & 21.80 
& 24.55 & 52.31 & 64.92 & 86.34 
& 59.54 & 79.88 & 89.69 \\

& & \methoditem{LDRE}{yang2024ldre}{SIGIR'24} & \cmark
& 23.35 & 24.03 & 26.44 & 27.50 
& 26.53 & 55.57 & 67.54 & 88.50 
& 60.43 & 80.31 & 89.90 \\

& & \methoditem{ImageScope}{luo2025imagescope}{WWW'25} & \cmark
& 25.39 & 25.82 & 27.07 & 27.98 
& 34.99 & 61.35 & 71.49 & 88.84 
& 74.94 & 88.24 & 94.02 \\

& & \methoditem{IP-CIR}{li2025imagine}{CVPR'25} & \cmark
& 26.43 & 27.41 & 29.87 & 31.07 
& 29.76 & 58.82 & 71.21 & 90.41 
& 62.48 & 81.64 & 90.89 \\

& & \methoditem{OSrCIR}{tang2025reason}{CVPR'25} & \cmark
& 23.87 & 25.33 & 27.84 & 28.97 
& 29.45 & 57.68 & 69.86 & -- 
& 62.12 & 81.92 & 91.10 \\

& & \methoditem{CoTMR}{sun2025cotmr}{ICCV'25} & \cmark
& 27.61 & 28.22 & 30.61 & 31.70 
& 35.02 & 64.75 & 76.18 & 92.51 
& 69.39 & 85.75 & 93.33 \\

\cline{2-15}

& \multirow{4}{*}{\centering Agentic}

& \methoditem{MRA-CIR}{tu2025multimodal}{ArXiv'25} & \xmark
& 27.14 & 28.85 & 31.54 & 32.63 
& 37.98 & 67.45 & 78.07 & 93.98 
& -- & -- & -- \\

& & \methoditem{AutoCIR}{cheng2025generative}{KDD'25} & \cmark
& 24.05 & 25.14 & 27.35 & 28.36 
& 31.81 & 61.95 & 73.86 & 92.07 
& 67.21 & 84.89 & 93.13 \\

& & \methoditem{$\mathrm{X}^{\mathrm{R}}$}{yang2026xr}{WWW'26} & \cmark
& \underline{31.38} & \underline{32.88} & \underline{35.46} & \underline{36.50} 
& \underline{43.13} & \underline{73.59} & \underline{83.09} & \underline{94.05} 
& \underline{77.98} & \best{90.68} & \underline{95.06} \\

& & \textbf{PDF (Ours)} & \cmark
& \best{43.04} & \best{43.73} & \best{46.03} & \best{46.86} 
& \best{43.65} & \best{74.65} & \best{83.53} & \best{95.25} 
& \best{78.74} & \underline{89.61} & \best{95.66} \\

\Xhline{0.8pt}

\end{tabular}
}
\end{table*}

\subsection{Memory-Guided Reasoning Refiner}
\textbf{Reasoning Experience Distillation.} In the candidate discrimination stage, LLMs often suffer from Logic Drift when performing multi-attribute cross-comparisons within long-context sequences. To mitigate this, we propose training-free Reasoning Experience Distillation (RED). We first construct an offline Experience Extraction dataset 
$\mathcal{D}_{\mathrm{ee}}$, where each instance $d \in \mathcal{D}_{ee}$ is represented as a five-tuple:
\begin{equation}
d = \left( \mathcal{C}(I_{ref}), T_{mod}, C_{pred}, \mathcal{C}_{k}, \mathcal{A}^* \right),
\end{equation}
where $\mathcal{C}_{k}=\{(id_j, C(I_j))\}_{j=1}^{k}$ denotes the set of identifiers $id_j$ and semantic proxies of the $k$ candidates. $\mathcal{A}^{*}\subseteq\{id_1,\ldots,id_K,\texttt{next\_page}\}$ denotes the predefined standard answer set, where $\texttt{next\_page}$ indicates that no correct answer exists among the candidates on the current page. To internalize robust heuristics, we design three sampling paradigms: \textit{Intra-page Truth Pursuit} for position-independent consistency, \textit{Cross-page Rejection Logic} for robust thresholding, and \textit{Counterfactual Distractor Defense} for fine-grained attribute sensitivity. Details of the construction of $\mathcal{D}_{\mathrm{ee}}$ are provided in the supplementary material. In the distillation phase, we draw inspiration from the group-relative comparison principle of GRPO and employ $\text{Logical\_Score}$ to assess 
the logical soundness of each reasoning chain. For each instance $d$, the base model $\pi_0$ generates $M$ reasoning rollouts $\mathcal{N} = \{ (think_z, y_z) \}_{z=1}^M$, with the quality quantified by a reward function:
\begin{equation}
r_z = R(y_z, \mathcal{A}^*) + \lambda \cdot \text{Logical\_Score}(think_z),
\end{equation}
where $R(\cdot)$ measures answer consistency. By contrasting superior and inferior outputs, we extract \textit{Semantic Advantages}:
\begin{equation}
\Delta_{sem} = \Gamma \left( d, \{ (y_z, r_z) \}_{z=1}^M, \mathcal{E}_t \right),
\end{equation}
where $\Gamma(\cdot)$ is an experience distiller. The library is then iteratively updated via an experience operator $\mathcal{U}(\cdot)$:
\begin{equation}
\mathcal{E}_{t+1} = \mathcal{U} \left( \mathcal{E}_t, \{ \Delta_{sem} \right),
\end{equation}
where $\mathcal{U}(\cdot)$ denotes the experience operator, which performs redundancy removal, conflict resolution, and condensation over the existing experiences. The resulting optimal external experience memory $\mathcal{E}^{*}$ is then solidified as the reasoning guideline for the DE-Manager in PDF.

\textbf{Tournament-based Test-Time Scaling.} We set the candidate pool size to $K=50$ and introduce the idea of Test-Time Scaling. Specifically, $\mathcal{B}_{50}$ is divided into two logically equivalent subsets, $S_1=\mathcal{B}_{1:25}$ and $S_2=\mathcal{B}_{26:50}$, enabling horizontal expansion of reasoning computation through a step-wise decision path.

We designed personalized Inference-compute Scaling paths based on dataset characteristics. For $|\mathcal{Y}_q|=1$, we employ a \textit{Sequential Staged Selection}. DE-Manager $\mathcal{D}(\cdot)$ first selects a local winner from $\mathcal{S}_1$:
\begin{equation}
I_{local}^{(1)} = \mathcal{D}(q, \mathcal{S}_1, \mathcal{E}^{*}),
\end{equation}
which is then injected into the context of $\mathcal{S}_2$ as the Leading Candidate to obtain the final answer:
\begin{equation}
I_{global} = \mathcal{D}(q, \mathcal{S}_2 \cup \{I_{local}^{(1)}\}, \mathcal{E}^{*}),
\end{equation}
For $|\mathcal{Y}_q|>1$, the model adopts \textit{Parallel Breadth Scanning}, performing independent screenings on $\mathcal{S}_1$ and $\mathcal{S}_2$ before merging results via a union operation $\mathcal{A}_{final} = \mathcal{A}_1 \cup \mathcal{A}_2$. In both cases, if no target is identified in the current subset, a \texttt{next\_page} instruction is output as feedback. The final decision result $y^*$ is expressed as:
\begin{equation}
y^* = \text{Refiner} \left( q, (\Phi(|\mathcal{Y}_q|)) \mid L, \mathcal{E}^{*} \right),
\end{equation}
where $\Phi(\cdot)$ is an adaptive strategy operator and $L=2$ denotes the reasoning stages. The unselected candidates in $\mathcal{B}_K$ are then ranked afterward in their original order. The value of $|\mathcal{Y}_q|$ is determined by the dataset protocol rather than query-specific ground-truth information, and therefore does not introduce ground-truth leakage.

\begin{table*}[t]
\centering

% 定义微缩会议号和 XR 上标宏，确保在 tiny 字号下比例协调
\newcommand{\conf}[1]{\hspace{1pt}\scalebox{0.75}{\textit{#1}}}
\newcommand{\XR}{X\textsuperscript{\scalebox{0.6}{R}}}

\caption{\textbf{Comparison with representative baselines on the Fashion IQ validation set.}
The best results are highlighted with \protect\best{red boxes}, while the second-best results are \underline{underlined}. ``--'' indicates results not reported in the original paper.}
\label{tab:Fashion IQ_comparison}
\tiny
\setlength{\tabcolsep}{2.1pt}
\renewcommand{\arraystretch}{1.0}
\resizebox{\textwidth}{!}{
\begin{tabular}{lllc|cc|cc|cc|cc}
\Xhline{0.5pt}

\multirow{3}{*}{\centering\textbf{Backbone}}
& \multirow{3}{*}{\centering\textbf{Type}}
& \multirow{3}{*}{\centering\textbf{Method}}
& \multirow{3}{*}{\makecell{\textbf{Training-}\\\textbf{free}}}
& \multicolumn{8}{c}{\textbf{Fashion IQ}} \\

\cline{5-12}

& & &
& \multicolumn{2}{c|}{\textbf{Shirts}}
& \multicolumn{2}{c|}{\textbf{Dresses}}
& \multicolumn{2}{c|}{\textbf{Tops \& tees}}
& \multicolumn{2}{c}{\textbf{Avg.}} \\

& & &
& R@10 & R@50
& R@10 & R@50
& R@10 & R@50
& R@10 & R@50 \\

\hline

\multirow{10}{*}{\centering ViT-B/32}

& \multirow{7}{*}{\centering\makecell{CIR-\\dedicated}}

& PALAVRA~\cite{cohen2022my} \conf{ECCV'22} & \xmark
& 21.49 & 37.05 & 17.25 & 35.94 & 20.55 & 38.76 & 19.76 & 37.25 \\

& & SEARLE~\cite{baldrati2023zero} \conf{ICCV'23} & \xmark
& 24.44 & 41.61 & 18.54 & 39.51 & 25.70 & 46.46 & 22.89 & 42.53 \\

\cline{3-12}

& & CIReVL~\cite{karthik2024vision} \conf{ICLR'24} & \cmark
& 28.36 & 47.84 & 25.29 & 46.36 & 31.21 & 53.85 & 28.29 & 49.35 \\

& & LDRE~\cite{yang2024ldre} \conf{SIGIR'24} & \cmark
& 27.38 & 46.27 & 19.97 & 41.84 & 27.07 & 48.78 & 24.81 & 45.63 \\

& & ImageScope~\cite{luo2025imagescope} \conf{WWW'25} & \cmark
& 24.29 & 37.49 & 18.00 & 35.20 & 24.99 & 41.41 & 22.42 & 38.03 \\

& & OSrCIR~\cite{tang2025reason} \conf{CVPR'25} & \cmark
& 31.16 & 51.13 & 29.35 & 50.37 & 36.51 & 58.71 & 32.34 & 53.40 \\

& & CoTMR~\cite{sun2025cotmr} \conf{ICCV'25} & \cmark
& 33.42 & 53.93 & \underline{31.09} & \underline{54.54} & 38.40 & 61.14 & 34.30 & 56.54 \\

\cline{2-12}

& \multirow{3}{*}{\centering\makecell{Agentic\\CIR}}
& AutoCIR~\cite{cheng2025generative} \conf{KDD'25} & \cmark
& 32.43 & 51.67 & 26.52 & 46.36 & 33.96 & 56.09 & 30.97 & 51.37 \\

& & \XR{}~\cite{yang2026xr} \conf{WWW'26} & \cmark
& \underline{36.06} & \underline{54.66} & 30.94 & 52.06 & \underline{42.99} & \underline{64.56} & \underline{36.66} & \underline{57.10} \\

& & \textbf{PDF (Ours)} & \cmark
& \best{39.47} & \best{58.64}
& \best{34.79} & \best{54.88}
& \best{43.87} & \best{65.83}
& \best{39.38} & \best{60.12} \\

\Xhline{0.5pt}

\multirow{13}{*}{\centering ViT-L/14}

& \multirow{9}{*}{\centering\makecell{CIR-\\dedicated}}

& Pic2Word~\cite{saito2023pic2word} \conf{CVPR'23} & \xmark
& 26.20 & 43.60 & 20.00 & 40.20 & 27.90 & 47.40 & 24.70 & 43.70 \\

& & SEARLE~\cite{baldrati2023zero} \conf{ICCV'23} & \xmark
& 26.89 & 45.58 & 20.48 & 43.13 & 29.32 & 49.97 & 25.56 & 46.23 \\

& & LinCIR~\cite{gu2024language} \conf{CVPR'24} & \xmark
& 29.10 & 46.81 & 20.92 & 42.44 & 28.81 & 50.18 & 26.28 & 46.49 \\

& & FTI4CIR~\cite{lin2024fine} \conf{SIGIR'24} & \xmark
& 31.35 & 50.59 & 24.49 & 47.84 & 32.43 & 54.21 & 29.42 & 50.88 \\

\cline{3-12}

& & CIReVL~\cite{karthik2024vision} \conf{ICLR'24} & \cmark
& 29.49 & 47.40 & 24.79 & 44.76 & 31.36 & 53.65 & 28.55 & 48.57 \\

& & LDRE~\cite{yang2024ldre} \conf{SIGIR'24} & \cmark
& 31.04 & 51.22 & 22.93 & 46.76 & 31.57 & 53.64 & 28.51 & 50.54 \\

& & ImageScope~\cite{luo2025imagescope} \conf{WWW'25} & \cmark
& 27.82 & 41.76 & 20.18 & 37.48 & 28.61 & 44.42 & 25.54 & 41.22 \\

& & OSrCIR~\cite{tang2025reason} \conf{CVPR'25} & \cmark
& 33.17 & 52.03 & 29.70 & 51.81 & 36.92 & 59.27 & 33.26 & 54.37 \\

& & CoTMR~\cite{sun2025cotmr} \conf{ICCV'25} & \cmark
& 35.43 & 54.91 & 31.18 & \underline{55.04} & 38.55 & 61.33 & 35.05 & 57.09 \\

\cline{2-12}

& \multirow{4}{*}{\centering\makecell{Agentic\\CIR}}
& MRA-CIR~\cite{tu2025multimodal} \conf{ArXiv'25} & \xmark
& \underline{40.43} & \underline{60.20}
& \underline{31.87} & 54.23
& 41.25 & 62.51
& \underline{37.85} & \underline{58.98} \\

& & AutoCIR~\cite{cheng2025generative} \conf{KDD'25} & \cmark
& 34.00 & 53.43 & 24.94 & 45.81 & 33.10 & 55.58 & 30.68 & 51.60 \\

& & \XR{}~\cite{yang2026xr} \conf{WWW'26} & \cmark
& 38.91 & 56.82 & 28.71 & 52.50 & \underline{43.91} & \underline{62.57} & 37.18 & 57.30 \\

& & \textbf{PDF (Ours)} & \cmark
& \best{41.07} & \best{60.43}
& \best{35.00} & \best{55.33}
& \best{44.10} & \best{64.76}
& \best{39.71} & \best{59.97} \\

\Xhline{0.5pt}
\end{tabular}
}
\end{table*}

\section{Experiments}
\subsection{Experimental Setup}
\noindent\textbf{Datasets and Evaluation Metrics.}
We evaluate our PDF on three representative CIR benchmark datasets: Fashion IQ~\cite{wu2021fashion} focuses on the fashion domain and contains three subcategories: Dresses, Shirts, and Tops \& Tees. 
We use Recall@$K$ ($K \in \{10, 50\}$) for these three categories, as well as their average values, as the evaluation metrics. 
CIRR~\cite{liu2021image} is the first natural-scene CIR dataset with subset retrieval. 
Following the standard protocol, we report Recall@$K$ ($K \in \{1, 5, 10, 50\}$) and Recall$_{\text{subset}}$@$K$ ($K \in \{1, 2, 3\}$). 
CIRCO~\cite{baldrati2023zero} is the first CIR dataset with multiple ground-truth annotations, and we adopt mean Average Precision (mAP@$K$) ($K \in \{5, 10, 25, 50\}$) as the evaluation metric.

\noindent\textbf{Implementation Details.}
We adopt the \texttt{ViT-B/32} and \texttt{ViT-L/14} CLIP~\cite{radford2021learning} models from OpenCLIP as the base retrievers. For all LLM invocations, we use GPT-4o-mini by default. The smoothing constant $\tau$ for weighted rank fusion is set to 60. The candidate pool size is set to $K=50$, which is divided into two reasoning stages ($L=2$). All experiments are conducted on a single NVIDIA RTX A6000 GPU. 

\noindent\textbf{Baselines.} To comprehensively evaluate the effectiveness of the proposed PDF framework, we compare it with three categories of representative baselines: Training-dependent Representation Approaches and Generative Training-free Baselines under CIR-dedicated Methods, as well as Strategic Multi-agent Architectures. More details are provided in the supplementary material.

\subsection{Results}

\begin{table*}[t]
\centering
\caption{Ablation on core components of PDF. Red numbers indicate relative improvement over the baseline in the first row.}
\label{tab:ablation_core_components}
\scriptsize
\setlength{\tabcolsep}{2.8pt}
\renewcommand{\arraystretch}{1.15}
\resizebox{\textwidth}{!}{
\begin{tabular}{ccc|c|c|cc!{\vrule width 0.8pt}cccc}
\Xhline{0.8pt}
\multicolumn{3}{c|}{\textbf{Multi-view Priors}}
& \multirow{2}{*}{\textbf{Delib.}}
& \multirow{2}{*}{\textbf{Fusion}}
& \multicolumn{2}{c!{\vrule width 0.8pt}}{\textbf{Fashion IQ Avg.}}
& \multicolumn{4}{c}{\textbf{CIRCO}} \\
\cline{1-3} \cline{6-11}
$W_{pred}$ & $W_{key}$ & $W_{vis}$
&  & 
& R@10 & R@50
& mAP@5 & mAP@10 & mAP@25 & mAP@50 \\
\Xhline{0.8pt}

\cmark & - & - 
& - & - 
& 31.82 & 53.40
& 25.18 & 26.02 & 28.54 & 29.50 \\

\cmark & \cmark & - 
& - & IPR
& 34.14 & 56.30
& 27.43 & 29.05 & 30.74 & 32.01 \\
\hline

\cmark & \cmark & \cmark
& - & Avg.
& 34.71 & 57.52
& 19.61 & 20.91 & 23.66 & 24.73 \\

\cmark & \cmark & \cmark
& - & Static
& 36.26 & 58.89
& 28.41 & 30.52 & 32.08 & 33.21 \\

\cmark & \cmark & \cmark
& - & IPR
& 37.57 & 60.20
& 30.78 & 31.79 & 34.21 & 35.35 \\
\hline

\cmark & \cmark & \cmark
& \cmark & IPR
& 39.38 {\color{red}(23.76\%)}
& 60.20 {\color{red}(12.73\%)}
& 40.02 {\color{red}(58.94\%)}
& 40.12 {\color{red}(54.19\%)}
& 41.75 {\color{red}(46.29\%)}
& 42.50 {\color{red}(44.07\%)} \\
\Xhline{0.8pt}
\end{tabular}
}
\end{table*}

\textbf{Overall Performance.}
Tables~\ref {tab:circo_cirr_comparison} and ~\ref {tab:Fashion IQ_comparison} demonstrate that PDF shows clear advantages on CIRCO and Fashion IQ. Compared with the strongest agentic baseline $X^{\mathrm{R}}$~\cite{yang2026xr}, PDF achieves a 45.47\% improvement in mAP@5 on CIRCO under ViT-B/32, and improves the average R@10 on Fashion IQ by 7.4\%. On CIRR, PDF also reaches state-of-the-art performance and outperforms $\mathrm{X}^{\mathrm{R}}$ on $78.57\%$ of the reported metrics. These results demonstrate the effectiveness of the proposed hierarchical reasoning paradigm in handling complex compositional intents.

\textbf{Training-dependent Methods.} PDF adaptively mobilizes three complementary priors through the IPR mechanism. Compared with FTI4CIR under ViT-L/14, PDF improves mAP@5 by $185.98\%$ on CIRCO, Recall@1 by $68.53\%$ on CIRR, and the average R@10 by $34.98\%$ on Fashion IQ. These results indicate that IPR exhibits stronger adaptability to semantic shifts than fixed training-based features.

\textbf{Generative Training-free Baselines.} PDF alleviates this limitation by dynamically integrating three complementary priors. Compared with CoTMR under ViT-B/32, PDF improves mAP@5 by $78.98\%$ on CIRCO and Recall@1 by $26.11\%$ on CIRR. These improvements demonstrate that the multi-view intent routing and fine-selection mechanism of PDF can better handle noisy real-world retrieval spaces, where relying solely on LLM-generated target descriptions may be insufficient.

\textbf{Strategic Multi-agent Architectures.}
Unlike AutoCIR, which is constrained by iterative redundancy and the upper bound of the retriever, and $X^R$, which relies on isolated verification under static orchestration, PDF enhances cross-candidate discrimination and rank correction through dynamic intent routing and experience-driven tournament-style TTS, thereby shifting the retrieval paradigm from perceptual matching toward deliberate logical decision-making without parameter updates.

\subsection{Ablation Study}
In this section, we conduct ablation experiments with ViT-B/32 on Fashion IQ and CIRCO to systematically analyze the key designs of PDF. Specifically, we study the effectiveness of the core components, the adaptiveness of intent-aware prior routing, the influence of deliberation depth and experience guidance, and the trade-off between candidate pool size and test-time computational cost.

Table~\ref{tab:ablation_core_components} validates the effectiveness of the core components in PDF. Under the same IPR strategy, the model performance consistently improves as $W_{key}$ and $W_{vis}$ are incorporated, indicating that multi-view perceptual priors provide complementary information. After introducing the Delib. stage, the model achieves the best performance. Compared with the baseline using only $W_{pred}$, it brings relative improvements of $23.76\%$ and $58.94\%$ on Fashion IQ R@10 and CIRCO mAP@5, respectively, demonstrating the fine-grained discriminative capability of deliberative reasoning. The comparison of fusion strategies further shows that dynamic IPR clearly outperforms average fusion and static fusion. In particular, average fusion leads to performance degradation on CIRCO, suggesting that dynamic resource scheduling is more suitable for heterogeneous CIR scenarios. In addition, we observe that the performance gains on Fashion IQ rely more on the IPR strategy, while CIRCO benefits more from the Delib. stage. We attribute this to the fact that dataset-level semantic determinacy affects the contribution of different PDF components. Detailed analysis is provided in the supplementary material.

\begin{table}[t]
\centering
\caption{Ablation study of deliberation depth and experience guidance with fixed candidate pool size $K=50$. Bold values indicate the best performance across different deliberation-depth settings.}
\label{tab:ablation_deliberation_depth}
\scriptsize
\setlength{\tabcolsep}{3.0pt}
\renewcommand{\arraystretch}{1.18}
\resizebox{\columnwidth}{!}{
\begin{tabular}{c|c|c|cccc}
\Xhline{0.8pt}
\multirow{2}{*}{\textbf{$L$}}
& \multirow{2}{*}{\textbf{Experience}}
& \textbf{Fashion IQ Avg.}
& \multicolumn{4}{c}{\textbf{CIRCO}} \\
\cline{3-7}
& 
& R@10
& mAP@5 & mAP@10 & mAP@25 & mAP@50 \\
\Xhline{0.6pt}

0 & -
& 37.57
& 30.78 & 31.79 & 34.21 & 35.35 \\
\hline

1 & -
& 36.97
& 35.04 & 35.99 & 37.47 & 38.83 \\

1 & \cmark
& 37.25
& 35.83 & 36.76 & 38.53 & 39.44 \\

2 & -
& 38.36
& 37.19 & 37.48 & 39.55 & 40.38 \\

2 & \cmark
& \textbf{39.38}
& \textbf{40.02} & \textbf{40.12} & \textbf{41.75} & \textbf{42.50} \\
\hline

3 & \cmark
& 38.47
& 38.22 & 38.93 & 40.06 & 40.72 \\
\Xhline{0.8pt}
\end{tabular}
}
\end{table}

Table~\ref{tab:ablation_deliberation_depth} analyzes the effects of reasoning depth $L$ and experience guidance under a fixed candidate pool size of $K=50$. As observed, after introducing the Delib. stage, the model achieves consistent performance improvements on CIRCO and obtains the best results when $L=2$, demonstrating the effectiveness and necessity of TTS. In contrast, the performance drops when $L=3$, suggesting that excessive reasoning depth may introduce redundant judgments and accumulated errors. Moreover, under the same $L$, incorporating experience guidance consistently outperforms the setting without experience, verifying that the decision experience derived from RED can effectively guide the model to perform structured verification. It is also worth noting that when $L=1$, the performance on Fashion IQ slightly decreases. We attribute this to the subjective nature of Fashion IQ descriptions, which may prevent the reasoning benefits from being fully realized.

\begin{figure}[t]
    \centering
    \includegraphics[width=\columnwidth]{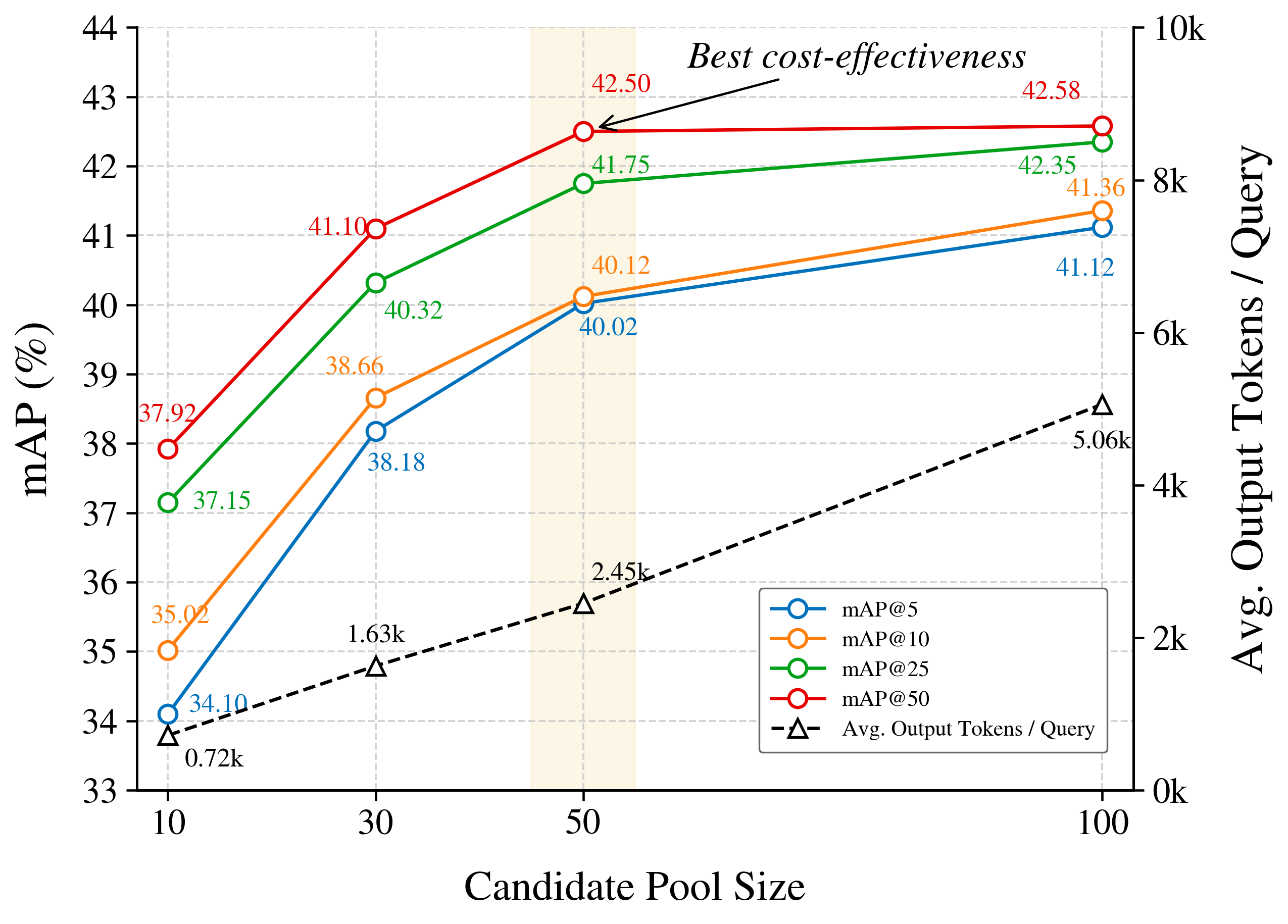}
    \caption{
    Effect of candidate pool size $K$ on retrieval performance (solid lines) and test-time average output tokens per
query (dashed line) on CIRCO. The shaded region highlights the optimal trade-off point at $K=50$.
    }
    \label{fig:candidate_pool_cost}
\end{figure}

Fig.~\ref{fig:candidate_pool_cost} analyzes the effect of the candidate pool size $K$ on retrieval performance and inference cost, where the inference cost is measured by the average number of output tokens per query. Under the setting of fixed $L=2$ with experience guidance, increasing $K$ from $10$ to $50$ leads to clear improvements across all mAP metrics, indicating that expanding the test-time reasoning space can effectively improve candidate recall quality and fine-grained semantic alignment. However, when $K$ is further increased to $100$, the performance gain becomes saturated, while the number of output tokens increases significantly, suggesting higher computational cost and diminishing returns. Considering both retrieval accuracy and inference efficiency, $K=50$ achieves the best trade-off and is therefore used as the default candidate pool size in this work. More visualizations and case analyses are provided in the supplementary material.

\section{Conclusion}
In this paper, we introduce a Perception-to-Deliberation Framework, a hierarchical framework composed of Intent-aware Prior Routing (IPR) and a Memory-Guided Reasoning Refiner (MRR): IPR dynamically orchestrates three complementary priors, including target imagination, explicit constraints, and visual anchors, effectively addressing the perceptual myopia and semantic drift caused by static fusion under diverse modification intents. Guided by offline-distilled experience, MRR further improves reasoning density through a Tournament-style Test-Time scaling strategy. Both components are training-free and plug-and-play, and achieve significant performance improvements on three widely used benchmarks, including CIRR, CIRCO, and Fashion IQ, fully validating the effectiveness of the ``thinking instead of searching'' paradigm.

% \section*{Acknowledgments}
% The authors thank the anonymous reviewers for their valuable comments and suggestions.

% \bibliographystyle{aaai2027}
\bibliography{AAAI-references}

% \clearpage
% \appendix
% \input{AAAI-appendix}


\end{document}